\documentclass[sigconf]{acmart}
\AtBeginDocument{%
  }

\usepackage{booktabs}
\usepackage{multirow}
\usepackage{amsmath}
\usepackage{tikz}
\usepackage{tcolorbox}
\usepackage{xcolor}
\usepackage{listings}
\usepackage[scaled=0.8]{beramono}
\usepackage{enumitem}

\lstdefinelanguage{penman}{
    morekeywords={:ent, :rol, :nam, :occ, :hco, :loc, :dob, :dom, :moy, :yoc, :pob, :dod, :pod, :geo, :equ, :tgt},
    sensitive=true,
}

\copyrightyear{2025}
\acmYear{2025}
\setcopyright{acmcopyright}
\setcctype{by}
\acmConference[MM '25]{Proceedings of the 33rd ACM International Conference on Multimedia}{October 27--31, 2025}{Dublin, Ireland}
\acmBooktitle{Proceedings of the 33rd ACM International Conference on Multimedia (MM '25), October 27--31, 2025, Dublin, Ireland}
\acmDOI{10.1145/3746027.3755444}
\acmISBN{979-8-4007-2035-2/2025/10}

\begin{document}

%%
%% The "title" command has an optional parameter,
%% allowing the author to define a "short title" to be used in page headers.
\title{Multi-Modal Semantic Parsing for the Interpretation of Tombstone Inscriptions}

%%
%% The "author" command and its associated commands are used to define
%% the authors and their affiliations.
%% Of note is the shared affiliation of the first two authors, and the
%% "authornote" and "authornotemark" commands
%% used to denote shared contribution to the research.
% \author{Ben Trovato}
% \authornote{Both authors contributed equally to this research.}
% \email{trovato@corporation.com}
% \orcid{1234-5678-9012}
% \author{G.K.M. Tobin}
% \authornotemark[1]
% \email{webmaster@marysville-ohio.com}
% \affiliation{%
%   \institution{Institute for Clarity in Documentation}
%   \city{Dublin}
%   \state{Ohio}
%   \country{USA}
% }

 \author{Xiao Zhang}
   \email{xiao.zhang@rug.nl}
\orcid{0000-0002-9582-7662}
 \affiliation{%
  \institution{University of Groningen}
  \city{Groningen}
  \country{Netherlands}
}

\author{Johan Bos}
  \email{johan.bos@rug.nl}
\orcid{0000-0002-9079-5438}
 \affiliation{%
  \institution{University of Groningen}
  \city{Groningen}
  \country{Netherlands}
}

% \author{Lars Th{\o}rv{\"a}ld}
% \affiliation{%
%   \institution{The Th{\o}rv{\"a}ld Group}
%   \city{Hekla}
%   \country{Iceland}}
% \email{larst@affiliation.org}

% \author{Valerie B\'eranger}
% \affiliation{%
%   \institution{Inria Paris-Rocquencourt}
%   \city{Rocquencourt}
%   \country{France}
% }

% \author{Aparna Patel}
% \affiliation{%
%  \institution{Rajiv Gandhi University}
%  \city{Doimukh}
%  \state{Arunachal Pradesh}
%  \country{India}}

% \author{Huifen Chan}
% \affiliation{%
%   \institution{Tsinghua University}
%   \city{Haidian Qu}
%   \state{Beijing Shi}
%   \country{China}}

% \author{Charles Palmer}
% \affiliation{%
%   \institution{Palmer Research Laboratories}
%   \city{San Antonio}
%   \state{Texas}
%   \country{USA}}
% \email{cpalmer@prl.com}

% \author{John Smith}
% \affiliation{%
%   \institution{The Th{\o}rv{\"a}ld Group}
%   \city{Hekla}
%   \country{Iceland}}
% \email{jsmith@affiliation.org}

% \author{Julius P. Kumquat}
% \affiliation{%
%   \institution{The Kumquat Consortium}
%   \city{New York}
%   \country{USA}}
% \email{jpkumquat@consortium.net}

%%
%% By default, the full list of authors will be used in the page
%% headers. Often, this list is too long, and will overlap
%% other information printed in the page headers. This command allows
%% the author to define a more concise list
%% of authors' names for this purpose.
\renewcommand{\shortauthors}{Xiao Zhang and Johan Bos}

%%
%% The abstract is a short summary of the work to be presented in the
%% article.
\begin{abstract}
Tombstones are historically and culturally rich artifacts, encapsulating individual lives, community memory, historical narratives and artistic expression. Yet, many tombstones today face significant preservation challenges, including physical erosion, vandalism, environmental degradation, and political shifts. In this paper, we introduce a novel multi-modal framework for tombstones digitization, aiming to improve the interpretation, organization and retrieval of tombstone content. Our approach leverages vision-language models (VLMs) to translate tombstone images into structured Tombstone Meaning Representations (TMRs), capturing both image and text information. To further enrich semantic parsing, we incorporate retrieval-augmented generation (RAG) for integrate externally dependent elements such as toponyms, occupation codes, and ontological concepts. Compared to traditional OCR-based pipelines, our method improves parsing accuracy from an F1 score of 36.1 to 89.5. Furthermore, we evaluate the model’s robustness across diverse linguistic and cultural inscriptions, and simulate physical degradation through image fusion to assess performance under noisy or damaged conditions. Our work represents the first attempt to formalize tombstone understanding using large vision-language models, presenting implications for heritage preservation. The code and supplementary materials are available at: \url{https://github.com/LastDance500/Tombstone-Parsing}.
\end{abstract}

%%
%% The code below is generated by the tool at http://dl.acm.org/ccs.cfm.
%% Please copy and paste the code instead of the example below.
%%


%%
%% Keywords. The author(s) should pick words that accurately describe
%% the work being presented. Separate the keywords with commas.
%% A "teaser" image appears between the author and affiliation
%% information and the body of the document, and typically spans the
%% page.
% \begin{teaserfigure}
%   \includegraphics[width=\textwidth]{sampleteaser}
%   \caption{Seattle Mariners at Spring Training, 2010.}
%   \Description{Enjoying the baseball game from the third-base
%   seats. Ichiro Suzuki preparing to bat.}
%   \label{fig:teaser}
% \end{teaserfigure}

\begin{teaserfigure}
    \centering
    \includegraphics[width=0.95\linewidth]{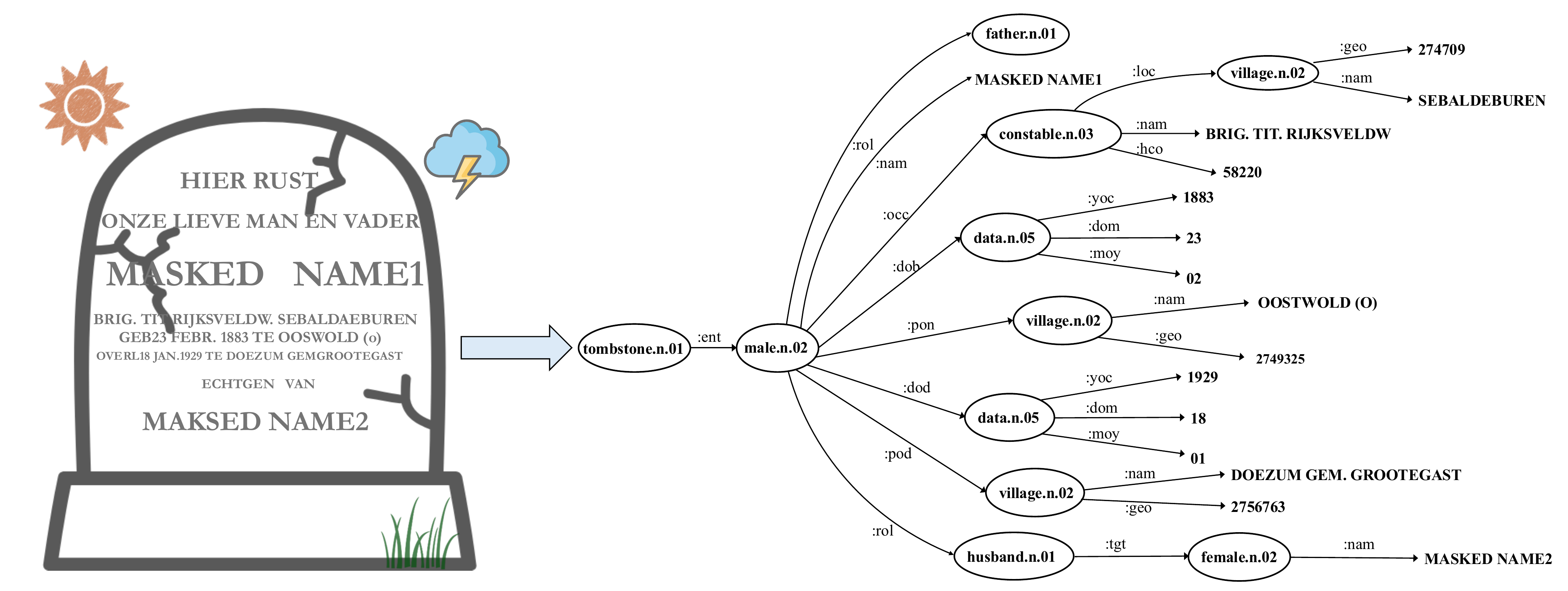}
    \caption{The overall idea of tombstone parsing. The input is an image of a tombstone, and the output is semantically interpreted, structured information, that can be stored in a relational database or semantic graph.}
    \label{fig:example}
\end{teaserfigure}

% \received{15 April 2025}
% % \received[revised]{12 March 2009}
% \received[accepted]{4 Jul 2025}

%%
%% This command processes the author and affiliation and title
%% information and builds the first part of the formatted document.
\maketitle

\section{Introduction}

Tombstone inscriptions give us rich historical and cultural insight. Therefore, cemeteries have sparked the interest of many different types of researchers, including anthropologists, archaeologists, genealogists, historians, philologists, sociologists, and theologists \cite{saller1984tombstones,eckert1993language,carmack:2002,veit2008new,long2016disambiguating,bos2022semantically,chen2024obi}. Besides the names of the deceased and dates and location of birth and death, tombstones often also contain information about family relationships, occupations, epitaphs, biblical quotes, poetry, and more. However, the intricate nature of these tombstones makes it challenging to interpret, organize and retrieve the valuable information they contain. Moreover, many tombstones are at risk due to physical erosion, vandalism, environmental degradation, and political shifts \cite{matthias1967weathering, streiter2007tombstones}. Digital preservation is therefore essential---not only to safeguard these irreplaceable cultural records but also to facilitate storage and search in structured databases.

With the ultimate aim of creating a large database with structured information of tombstone inscriptions, we investigate the possibility of automatically parsing tombstone images into structured data suitable for storage, in order to assist data curators and provide cemetery researchers with a powerful tool for searching into databases of digitized tombstones. The overall idea can be seen in Figure~\ref{fig:example}, where the input is an image (photograph) of a tombstone, and the output is an entry in a relational database.\footnote{Names of people on tombstones shown in this paper have been masked. Our aim is to stimulate research in developing automatic tools for cemetery research rather than distributing private information online.} Cemetery researchers are then able to systematically search tombstones on dates, toponyms, gender, jobs, relationships, and other features.

Our work has points of contact with the XML-based annotation scheme for graveyards and tombstones proposed by \citet{Streiter:2007}. However, their scheme focuses on the physical properties of tombstones, whereas we aim to analyse the content of tombstone inscriptions.  This content includes people, dates, locations, symbols, religious references that are mentioned on tombstones and the relations between them (temporal relations, family relations, occupational relations). The research by \citet{franken}, who developed a system for automatically recognising and interpreting ancient Egyptian hieroglyphs from photographs based on a dataset of about 4,000 hieroglyphs, is similar in objectives to our research, and so is other research in this area \cite{Zhang2024Ancient, assael2022restoring, jiang2024oraclesage}. But even closer to our goal is earlier work by
\citet{bos2022automatically}, who propose a pipeline-based approach consisting of image segmentation, OCR-based text extraction, and structured parsing. Their approach requires extensive manual annotation of relevant image segments, and has proven to be limited in effectiveness, achieving F1 scores below 40\%, even for a simplified version of the task. As a result, it is not scalable to real-world scenarios. Thus, in this paper, we investigate whether we can take advantage of recent advances in end-to-end vision-language models for this task.

Vision-Language Models (VLMs) \cite{openai2024gpt4,anthropic2024claude,geminiteam2024geminifamilyhighlycapable,bai2025qwen25vltechnicalreport} have demonstrated remarkable capabilities in image and text comprehension, enabling direct generation of natural language descriptions from images. This presents an opportunity to improve tombstone parsing by directly generating structured content from images of gravestones rather than relying on OCR-based methods in complex and vulnerable pipeline architectures. However, VLMs lack domain-specific knowledge---they do not inherently interpret human-defined identifiers, which are essentially arbitrary labels lack of intrinsic meaning, such as geocodes (location identifiers for toponyms), HISCO codes (occupational classification) and WordNet sense number (grounded meaning of a word). Drawing inspiration from prior work on retrieval-augmented generation (RAG) for formal textual semantic parsing \cite{zhang2024RASP}, we explore the way to incorporate RAG into the VLM pipeline to enable the retrieval and integration of relevant external knowledge. To this end, we propose the \textbf{Tomb2Meaning (T2M)} framework, which combines VLMs and RAG to convert tombstone images into formal, structured meaning representations. Our investigation is guided by the following three research questions in the context of automated tombstone parsing:

\begin{itemize}
    \item Can VLMs be leveraged to enhance tombstone parsing, outperforming previous OCR-based methods?\\[-6pt]
    
    \item Can RAG bridge the knowledge gap in neural tombstone parsing?\\[-6pt]
    
    \item How do models handle challenging stones, such as those with linguistically complex inscriptions or physical defects?
\end{itemize}

Following a review of background and related work on semantic parsing and tombstone meaning representations (Section~\ref{sec:related}), we introduce the T2M framework (Section~\ref{sec:methods}). We then present the experiments (Section~\ref{sec:exp}) that evaluate VLMs and RAG under few-shot and fine-tuned settings, including three strategies for integrating retrieval components. Our results (Section~\ref{sec:results}) show that fine-tuned VLMs achieve the state-of-the-art performance in tombstone parsing, with RAG further boosting accuracy by incorporating external domain knowledge. In Section~\ref{sec:challenges}, we assess the models on linguistically challenging inscriptions across five dimensions, and evaluate robustness under simulated physical degradation.

\section{Background and Related Work\label{sec:related}}

\subsection{Automated Formal Semantic Parsing}

Formal meaning representations play a crucial role in advancing natural language understanding (NLP) by providing structured, abstract depictions of sentence meaning. Abstract Meaning Representation (AMR) \cite{banarescu2013abstract} utilizes directed acyclic graphs to encapsulate the core semantic relationships within a sentence. Discourse Representation Structure (DRS) \cite{Kamp1993FromDT, gmb, Bos2023IWCS} is designed to capture more context-dependent phenomena such as anaphora, temporal expressions and quantification across extended discourse.  There are many other semantic formalisms, such as BMR \cite{martinez-lorenzo-etal-2022}, PTG \cite{hajic-etal-2012-announcing}, and EDS \cite{oepen-lonning-2006-discriminant}, focusing on different semantic phenomena. 

Semantic parsing is a cornerstone task in NLP that focuses on converting natural language text into structured formal meaning representations. This transformation involves mapping complex linguistic input into abstract structures---such as AMR, DRS, and others. Traditionally, this process has been tackled using rule-based systems \cite{woods1973progress,Hendrix1977DevelopingAN,templeton-burger-1983-problems} as well as neural models \cite{barzdins-gosko-2016-riga,noord-2017-neural,van-noord-etal-2018-exploring,van-noord-etal-2020-character,Bevilacqua2021OneST,wang-etal-2023-discourse, ettinger-etal-2023-expert, liu-2024-model, liu-2024-soft, yang2024DRSparsing, zhang-etal-2024-gaining, Zhang2024nerualsp}, primarily with textual input.

Recent advancements have extended semantic parsing into the multimodal domain, integrating non-linguistic modalities such as vision, speech or gesture. For instance, \citet{Abdelsalam2022VisualSP} introduced visual semantic parsing, translating scene graphs extracted from images into AMR-like representations. \citet{brutti-etal-2022-abstract} proposed gesture meaning representation, capturing the semantic content of hand gestures. Multimodal AMR parsing has also been explored, combining visual and textual inputs for enhanced grounding \cite{bonial-etal-2023-abstract,lai-etal-2024-encoding}. There are also many works in the field of Robotics. \citet{Thomason2020JointlyIP} used multi-modal learning to improve perceptual concept parsing for human-robot interaction. \citet{Farazi2020AttentionGS} introduced a semantic relationship parser for visual question answering, which performs multi-modal semantic parsing by generating semantic feature vectors from images and questions.

These works demonstrate the potential of extending formal semantic parsing to complex multimodal settings. However, the work which investigated multimodal semantic parsing in the context of historical visual artifacts are still under poor-exploration.

\subsection{Tombstone Meaning Representations}

Tombstone inscriptions convey rich relational information that goes beyond simple attribute–value pairs. They often include relationships between people (for instance spouse or husband) or a series of family relations (e.g., a person being involved in various marriages over time), which require expressive representational power. Logical attributes such as negation or universal quantification are rarely encountered on tombstone inscriptions. Hence, a suitable formalism is a rooted directed acyclic graph.

Tombstone Meaning Representations (TMRs) were introduced by \citet{bos2022automatically}, employing the PENMAN notation to encode directed acyclic graphs \cite{goodman-2020-penman} in a text-based notation where nodes are specified by a unique identifier, and labelled edges connect nodes to other nodes. PENMAN was originally developed for natural language generation systems \cite{Kasper:1989,Bateman:1989,BatemanParis:1989,Bateman:1990} and later also for natural language understanding \cite{noord-2017-neural,cai-knight-2013-smatch}. It is a bracketed structure of a directed acyclic graph, where nodes are identified with (unique) variable names. A TMR is defined by the following grammar:

\begin{quote}\footnotesize
\begin{verbatim}
TMR ::= "(" VAR " / " CCT ")" | "(" VAR " / " CCT RLS ")"
CCT ::= LEM "." POS "." SNS
RLS ::= REL TMR | REL DAT | REL LIT | REL VAR | RLS 
REL ::= " :ent " | " :nam " | " :txt " | ... | " :pod "
POS ::= "n" | "v" | "a" | "r"
SNS ::= "01" | "02" | ... | "99"
\end{verbatim}
\end{quote}

Variables names are written with one lowercase letter followed by a number.
Variable instances are written with a forward slash.
Concepts are represented as WordNet synsets \cite{Fellbaum-1998-wordnet}.
Relations are written using a colon followed by three lowercase letters (see Table~\ref{tab:relations}). Literals are used for date expressions, toponym identifiers (following the GeoNames geographical database) and HISCO \cite{van2004creating} historical international standard codes for occupations. Relations can be inversed---inversed roles have the suffix \texttt{-of}, and for any relation $R$ the following holds $R(x,y) \equiv R\texttt{-of}(y,x)$. An example TMR (corresponding to Figure~\ref{fig:example}) in PENMAN is provided below. %in Figure~\ref{fig:example}. 

\begin{lstlisting}
(t00000 / tombstone.n.01
    :ent (x1 / male.n.02
        :nam "Masked Name 1"
        :...
        :occ (x4 / constable.n.03
            :nam "BRIG. TIT. RIJKSVELDW."
            :hco "58220"
            :loc (x5 / village.n.02
                    :nam "SEBALDEBUREN"
                    :geo "2747409")))
        :...
        :dob (x6 / date.n.05
                :dom "23"
                :moy "02"
                :yoc "1883")   
        :...
        :rol (x10 / husband.n.01
                :tgt (x11 / female.n.02
                        :nam "Masked Name 2"))))
\end{lstlisting}

% \begin{figure}
%     \centering
%     \includegraphics[width=\linewidth]{images/TMR.pdf}
%     \caption{Tombstone and Tombstone Meaning Representation}
%     \label{fig:tmr}
% \end{figure}

\begin{table}
\caption{Binary relations with their interpretation used in Tombstone Meaning Representations}\label{tab:relations}
\small
\begin{tabular}{|c|l|}
\hline
\textbf{Relation} & \textbf{Explanation}\\
\hline
\texttt{:ent} & deceased entity mentioned on a stone\\
\texttt{:dob} & date of birth of a person\\
\texttt{:dod} & date of death of a person\\
\texttt{:pob} & place of birth of a person\\
\texttt{:pod} & place of death of a person\\
\texttt{:yoc} & year of century\\
\texttt{:moy} & month of year\\
\texttt{:dom} & day of month\\
\texttt{:equ} & equality link marking co-reference \\
\texttt{:pfx} & prefix of a name\\
\texttt{:sfx} & suffix of a name\\
\texttt{:beg} & begin of a period\\
\texttt{:end} & end of a period\\
\texttt{:rol} & role from a person with respect to another person\\
\texttt{:tgt} & The target person of a role\\
\texttt{:geo} & geographical reference, using \url{www.geonames.org}\\
\texttt{:occ} & occupation of a person\\
\texttt{:aft} & fulfilled in a time period after another role\\
\texttt{:bef} & fulfilled in a time period before another role\\
\hline
\end{tabular}
\end{table}

\section{Methodology}\label{sec:methods}

In this section, we first introduce three baselines for tombstone parsing. Then we describe our proposed framework, \textbf{Tomb2Meaning (T2M)}, with detailed discussion on the integration of retrieval-augmentation techniques at different stages in the pipeline.

\begin{figure*}
    \centering
    \includegraphics[width=\linewidth]{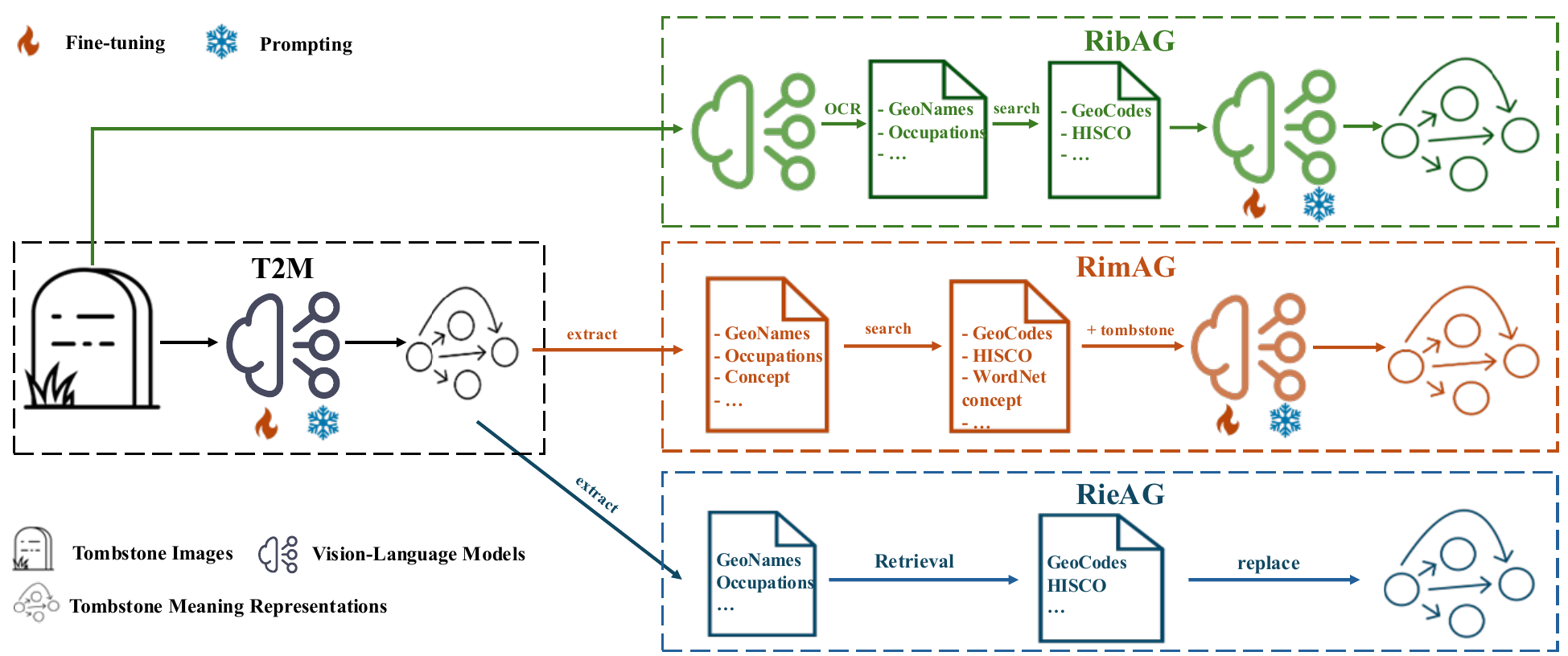}
    \caption{Overview of the Tomb2Meaning (T2M) framework. The three retrieval-augmented generation (RAG) strategies are: Retrieval-in-the-Beginning Augmented Generation (RibAG), Retrieval-in-the-Middle Augmented Generation (RimAG), and Retrieval-in-the-End Augmented Generation (RieAG). We denote two modes of model usage: direct prompting (indicated by the snowflake symbol) and fine-tuning (indicated by the flame symbol). Prompt templates and training settings are in supplemental materials.}
    \label{fig:pipeline}
\end{figure*}

\subsection{Baselines}

We consider three baselines for performance comparison. The first baseline replicates the multi-step pipeline from \citet{bos2022automatically}. Specifically, a fine-tuned YOLOv5 is employed for label detection to identify and classify relevant regions on the tombstone. An Optical Character Recognition (OCR) module then transcribes the textual content from these regions. Subsequently, post-OCR processing verifies locations, corrects names, and normalizes date expressions, followed by semantic interpretation to establish relationships among the extracted entities. We refer to this baseline as \textbf{YOLO-OCR}.

The second baseline streamlines the above process by prompting a Vision-Language Model (VLM) to extract the inscription directly from the image, thereby performing a VLM-based OCR. A subsequent fine-tuned inscription-to-TMR language model generates the Tombstone Meaning Representation (TMR). This approach is denoted as \textbf{VLM-OCR}.

To establish a low bound of performance, we introduce the deterministic, input-agnostic baseline parser called \textbf{Deterministic Baseline} \cite{may-2016-semeval}. It has been created by computing the average Smatch score for each TMR across training pairs. The TMR\footnote{This deterministic TMR can be seen in the supplemental materials.} with the highest average score is selected as a constant output, representing the minimal performance achievable by any well-designed system.

\subsection{The Tomb2Meaning framework}

The T2M framework integrates a VLM-based system with three RAG strategies.

\paragraph{\textbf{Base T2M}} 
Given a tombstone image \(I\), the base VLM \(f(\cdot)\) produces a TMR:

\begin{equation}
\mathrm{TMR}_{\text{base}} = f(I).
\end{equation}

\paragraph{\textbf{Retrieval Module \(R(\cdot)\) Implementation}}  
We define a retrieval component \(R(\cdot)\). Its implementation involves:
\begin{enumerate}
    \item Query Construction: Extracted entities \(E(\cdot)\), derived from either the input image or an initially generated TMR, are used to construct retrieval queries, which are then sent to the APIs of GeoNames, HISCO, and WordNet.
    \item Result Filtering: Retrieved results are filtered based on specific metrics. For GeoNames codes, we leverage the GPS coordinates of the tombstone images to select the geographically closest match. For WordNet synsets, any synset that does not correspond to the target part-of-speech is omitted.
    \item Prompt/Instruction Reconstruction: Once the external retrieval information is obtained, the prompts (or fine-tuning instructions) are reconstructed to incorporate this supplementary data.
\end{enumerate}

\paragraph{\textbf{RibAG (Retrieval-in-the-Beginning Augmented Generation)}}
In the RibAG, external retrieval is applied at the initial stage. The VLM is first prompted to extract key textual entities (e.g., place names, occupations) directly from the image \(I\). These extracted entities, \(E(I)\), are used to query external resources via \(R(\cdot)\). The retrieved information is then integrated into the generation process:

\begin{equation}
\mathrm{TMR}_{\text{RibAG}} = f\Big(I,\, R\big(E(I)\big)\Big)
\end{equation}

\paragraph{\textbf{RimAG (Retrieval-in-the-Middle Augmented Generation)}}
The RimAG strategy begins with generating an initial TMR:
\begin{equation}
\mathrm{TMR}_0 = f(I).
\label{equ:inital_tmr}
\end{equation}

Entities \(E(\mathrm{TMR}_0)\) are then extracted, and external codes are retrieved via \(R\big(E(\mathrm{TMR}_0)\big)\). The enhanced prompts/instructions that incorporate the retrieved data are fed back into the VLM:

\begin{equation}
\mathrm{TMR}_{\text{RimAG}} = f\Big(I,\, R\big(E(\mathrm{TMR}_0)\big)\Big).
\end{equation}

Unlike RibAG, RimAG leverages the initial TMR, which already includes potential WordNet concepts, enabling to retrieve not only GeoCodes, HISCO codes, but also WordNet synsets.

\paragraph{\textbf{RieAG (Retrieval-in-the-End Augmented Generation)}}
For the RieAG, after generating the initial TMR following Equation~\ref{equ:inital_tmr}. Entities \(E(\mathrm{TMR}_0)\) are extracted and external codes are retrieved. Instead of additional prompting or fine-tuning, the retrieved external information directly replaces corresponding sections in \(\mathrm{TMR}_0\) based on a predefined replacement strategy:

\begin{equation}
\mathrm{TMR}_{\text{RieAG}} = \operatorname{Replace}\Big(\mathrm{TMR}_0,\, R\big(E(\mathrm{TMR}_0)\big)\Big),
\end{equation}

where \(\operatorname{Replace}(\cdot)\) denotes the function that aligns and integrates external codes into the TMR, which can be achieved through a simple regular expression.

\section{Experiments\label{sec:exp}}

\subsection{Dataset}

Our dataset comprises 1,200 high-resolution photographs of tombstones, each manually annotated with a formal meaning representation following the TMR conventions.\footnote{https://www.let.rug.nl/bos/tombreader/} The images, stored in JPEG format, include metadata such as geolocation coordinates, timestamps, and date information. To ensure visual consistency, images were manually reoriented and cropped to remove extraneous background where needed.

While most inscriptions are in Dutch and rendered in serif fonts, the dataset also includes samples in English, French, German, Spanish, Italian, Indonesian, and Greek. Gothic script is another frequently occurring typeface. A small number of inscriptions exhibit multilingual mixing.

For experimental purposes, we have randomly divided these images into equal-sized training and testing sets (50:50 split), ensuring a sufficiently large sample for testing. Example images, annotations and dataset statistics are provided in our supplemental materials.

\subsection{Evaluation}

We adopt multiple complementary metrics to evaluate the accuracy and structural quality of the generated semantic graphs.

\paragraph{Smatch}  
Smatch \cite{cai-knight-2013-smatch} measures similarity between predicted and reference semantic graphs by converting each into a set of triples and computing an optimal variable mapping via hill-climbing. The precision (P), recall (R), and F1 score are computed as:
\begin{equation}
\begin{aligned}
    \text{P} &= \frac{m}{p}, \quad \text{R} = \frac{m}{g}, \quad \text{F1} = \frac{2 \cdot \text{P} \cdot \text{R}}{\text{P} + \text{R}},
\end{aligned}
\end{equation}
where \( m \) is the number of matched triples, \( p \) is the number of predicted triples, and \( g \) is the number of gold-standard triples.

\paragraph{Micro F1}  

To enable fine-grained evaluation of content accuracy, particularly for structured fields such as names, relations, date values, geographic identifiers, occupational codes (HISCO), and WordNet synsets, we compute the micro-averaged F1 score:

\begin{equation}
    \text{Micro F1} = \frac{2\,TP}{2\,TP + FP + FN}
\end{equation}

This metric aggregates true positives (TP), false positives (FP), and false negatives (FN) across all entity and relation types.

\paragraph{Ill-Formed Rate (IFR)}

To assess the structural correctness of the generated graphs, we introduce the Ill-Formed Rate (IFR). A graph is considered ill-formed if it contains structural issues such as cyclic dependencies, isolated nodes, or dangling edges pointing to non-existent elements. Such graphs are assigned a Smatch score and an F1 score of 0, providing a measure of structural errors.

\subsection{Models and Settings}

We evaluate our approaches using several state-of-the-art Vision-Language Models (VLMs), including LLava-v1.6-mistral-7B, Llama-3.2-vision-11B, and the Qwen2.5-VL series (3B, 7B, and 72B).

We conducted two experiments on the base framework: one using a frozen VLM via prompting and the other by fine-tuning the VLM. In the prompting experiment, we compared model performance using few-shot examples that were randomly selected from the train set. For the fine-tuning, we applied parameter-efficient fine-tuning (PEFT), specifically using LoRA\cite{hu2022lora}. All experiments are conducted on a standardized hardware setup, with detailed configurations provided in supplemental materials to support replication and future comparative studies.

% All experiments are conducted on a standardized hardware setup, with detailed configurations provided in Appendix~\ref{app:exp} to support replication and future comparative studies.\footnote{The code will be released after the anonymous review period.}

% In the LoRA method, the update for a pre-trained weight matrix \(\mathbf{W}_0 \in R^{d \times k}\) is represented as a low-rank decomposition:
% \begin{equation}
%     \Delta \mathbf{W} = \mathbf{B}\mathbf{A},
% \end{equation}

% where \(\mathbf{B} \in R^{d \times r}\), \(\mathbf{A} \in R^{r \times k}\), and \(r \ll \min(d,k)\). The adapted weight matrix is then:
% \begin{equation}
%     \mathbf{W} = \mathbf{W}_0 + \frac{\alpha}{r}\mathbf{B}\mathbf{A}.
% \end{equation}

% with \(\alpha\) as a scaling factor. Particularly, our fine-tuning includes all the layers of VLM, i.e., vision layers, language layers, attention modules and mlp modules.

\section{Results and Analysis\label{sec:results}}

\subsection{Few-shot Prompting}

\begin{figure}
    \centering
    \includegraphics[width=\linewidth]{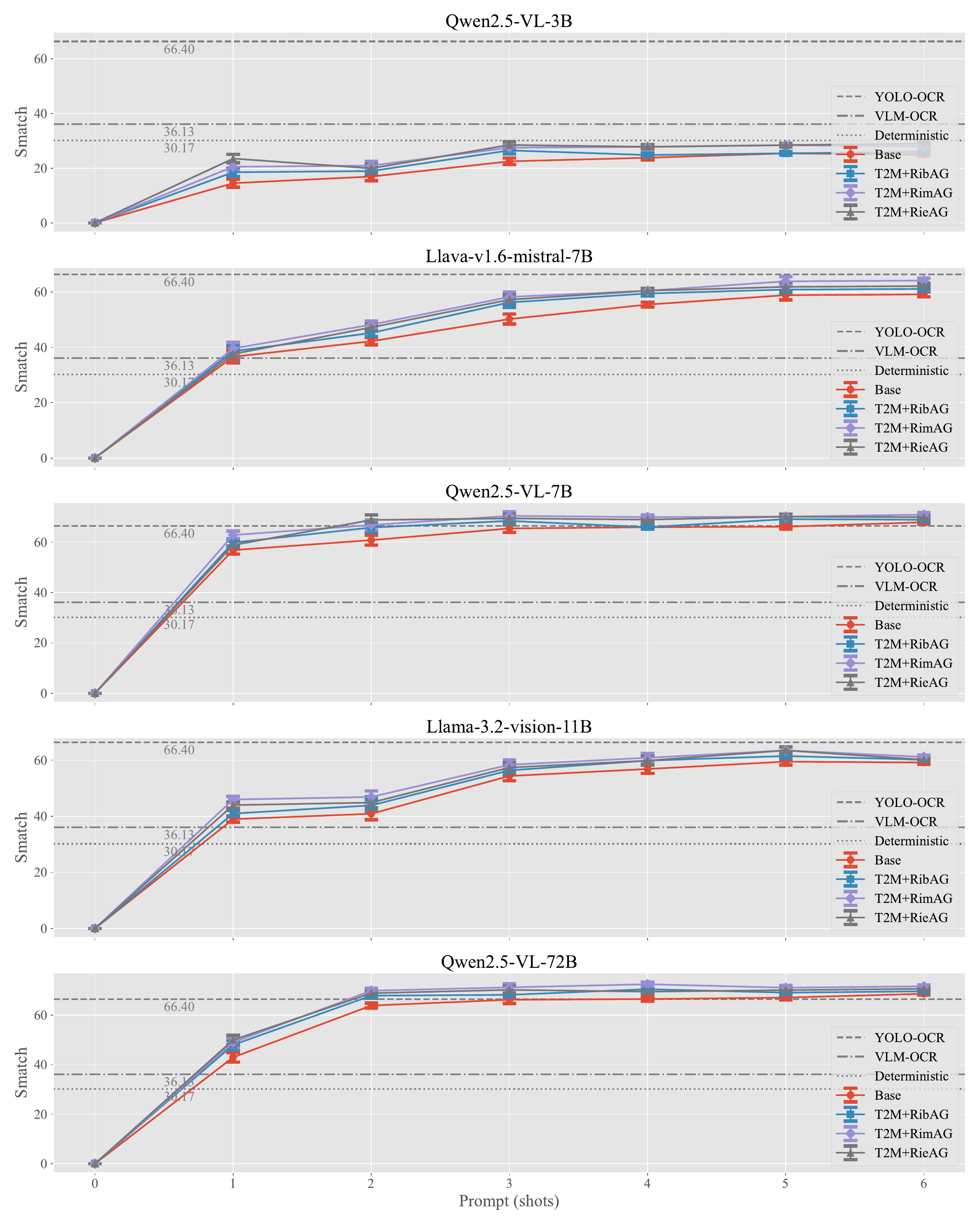}
    \caption{Smatch scores for five models (solid lines) and three baselines (dashed lines) under few-shot prompting, with the number of shots ranging from 0 to 6.}
    \label{fig:few_shot}
\end{figure}

\begin{table*}[htbp]
\centering
\caption{Performance of the automated tombstone parsing framework (T2M) with and without retrieval augmented generation methods for various models, compared to three baselines. Here, \emph{Smatch} is the overall F1-score, \emph{F1-X} represents the F1-score for instances of X, and \emph{IFR} is the ill-formed frequency rate of the parser output. All numbers are presented as percentages (\%).}
\label{tab:fine-tune}
\begin{tabular}{llcccccccc}
\toprule
\multirow{2}{*}{\textbf{Model}} & \multirow{2}{*}{\textbf{Method}} & \multirow{2}{*}{\textbf{Smatch}} & \multicolumn{3}{c}{\textbf{Internal}} & \multicolumn{3}{c}{\textbf{External}} & \multirow{2}{*}{\textbf{IFR}} \\
\cmidrule(lr){4-6} \cmidrule(lr){7-9}               &                       &                 & \textbf{F1-Name} & \textbf{F1-Role} & \textbf{F1-Date} & \textbf{F1-Geo} & \textbf{F1-Hisco} & \textbf{F1-Synset} & \\ \midrule
Deterministic Baseline & Baseline  & 30.17 & 25.81 & 28.69 & 30.10 & 00.00 & 00.00 & 32.10 & 0.00 \\ \midrule
YOLO-OCR~\cite{bos2022automatically} & Baseline  & 36.13 & 31.57 & 32.88 & 35.97 & 00.00 & 00.00 & 38.22 & 0.00 \\ \midrule
VLM-OCR & Baseline   & 66.40 & 65.17 & 65.83 & 64.80 & 00.23 & 01.58 & 75.21 & 3.16 \\ \midrule
\multirow{4}{*}{Llava-v1.6-mistral-7b} 
  & T2M   & 78.70 & 73.49 & 71.86 & 75.26 & 13.30 & 02.53 & 80.14 & 10.0 \\
  & T2M+RibAG  & 80.07 & 77.28 & 79.83 & 76.04 & 51.32 & 20.00 & 81.78 & 0.50 \\
  & T2M+RimAG  & 88.07 & 87.08 & 88.77 & 86.49 & \textbf{75.97} & \textbf{71.70} & 87.92 & 0.50 \\
  & T2M+RieAG  & 80.55 & 73.49 & 71.86 & 75.26 & 51.66 & 18.00 & 80.14 & 10.0 \\ \midrule
\multirow{4}{*}{Llama-3.2-vision-11B}  
  & T2M   & 80.90 & 71.90 & 81.27 & 83.38 & 15.99 & 00.00 & 80.07 & 1.00 \\
  & T2M+RibAG  & 82.04 & 76.40 & 79.37 & 84.15 & 44.97 & 18.18 & 81.13 & 0.50 \\
  & T2M+RimAG  & 85.88 & 79.99 & 84.08 & 85.81 & 59.66 & 59.85 & 83.94 & 0.60 \\
  & T2M+RieAG  & 83.20 & 71.90 & 81.27 & 83.38 & 58.54 & 15.74 & 80.07 & 1.00 \\ \midrule
\multirow{4}{*}{Qwen2.5-vl-3B} 
  & T2M   & 82.06 & 77.24 & 88.45 & 86.96 & 00.00 & 00.00 & 84.41  & 3.10 \\
  & T2M+RibAG  & 84.29 & 79.22 & 86.86 & 86.93 & 38.47 & 27.37 & 85.61 & 2.17 \\
  & T2M+RimAG  & 86.80 & 81.62 & 88.35 & 87.44 & 50.57 & 43.75 & 86.81 & 1.67 \\
  & T2M+RieAG  & 84.38 & 77.24 & 88.45 & 86.96 & 48.50 & 40.22 & 84.41 & 3.10 \\ \midrule
\multirow{4}{*}{Qwen2.5-vl-7B} 
  & T2M   & 85.80 & 82.38 & 90.99 & 89.58 & 10.83 & 00.00 & 88.12 & 2.17 \\
  & T2M+RibAG  & 88.13 & 82.12 & 91.27 & 88.80 & 43.45 & 21.82 & 88.63 & 0.83 \\
  & T2M+RimAG  & \textbf{89.50} & \textbf{84.83} & \textbf{91.32} & \textbf{89.94} & 66.58 & 60.33 & \textbf{90.52} & 2.17 \\
  & T2M+RieAG  & 88.54 & 82.38 & 90.99 & 89.58 & 69.57 & 16.00 & 88.12 & 2.17 \\ \bottomrule
\end{tabular}
\end{table*}

Figure~\ref{fig:few_shot} illustrates the impact of increasing prompt examples (from 0 to 6 shots) across five vision-language models. We observe several consistent trends: First, increasing the number of shots yields substantial performance gains in the early stages---particularly from zero-shot to one-shot---after which improvements taper off, indicating a saturation effect. Second, model size correlates with few-shot learning ability: larger models such as Qwen2.5-VL-7B show sustained improvement and eventually surpass the VLM-OCR baseline. In contrast, smaller models (e.g., Qwen2.5-VL-3B) exhibit only modest and sometimes unstable gains. Third, even with one-shot prompting, most VLMs already outperform the deterministic and OCR-based baselines, highlighting the advantages of general-purpose vision-language models.

Finally, the incorporation of RAG techniques consistently enhances performance across all models. In particular, RimAG shows the most substantial improvements, likely due to its ability to ground external references—such as GeoNames identifiers, HISCO codes, and WordNet synsets—via retrieved contextual information. These results suggest that while VLMs are already strong semantic parsers, their performance can be significantly enhanced when equipped with task-specific external knowledge.

\subsection{Fine-Tuning}

We now evaluate the impact of fine-tuning on tombstone parsing performance. Table~\ref{tab:fine-tune} compares results for the base T2M framework and its three RAG-enhanced variants across four VLMs.

Fine-tuning yields a notable performance boost over few-shot prompting, with Smatch scores improving by more than 10 percentage points---rising from the 70s to the low-to-mid 80s depending on model size. Among the retrieval-augmented variants, RimAG consistently outperforms the others, with peak Smatch performance reaching 89.50\% on Qwen2.5-VL-7B. RibAG provides moderate gains, while RieAG performs well but shows more variation across models.

To better understand performance differences, we analyze both internal and external fields. Internal fields, such as names, roles, and date structures, benefit from fine-tuning and RAG. In contrast, external fields, like GeoCodes, HISCO codes, and WordNet synsets, require knowledge beyond the input image, and therefore show the largest improvements when using RimAG. Notably, RimAG also reduces the ill-formed rate (IFR), indicating improvements in both semantic accuracy and structural integrity.

While RAG techniques are primarily designed to support external knowledge integration, we find that even internal field accuracy improves with retrieval-enhanced methods. This suggests that contextual augmentation during training helps VLMs better align visual content with structured meaning representations.

Overall, these results demonstrate that both fine-tuning and retrieval augmentation are critical to achieving high-quality, structurally well-formed TMRs in automated tombstone parsing.

\section{Evaluating Specific Challenges}\label{sec:challenges}

In this section, we examine the main challenges associated with tombstone parsing and evaluate the effectiveness of our framework in handling both linguistic and visual complexity.

\begin{figure*}[h]
    \centering
    \includegraphics[width=\linewidth]{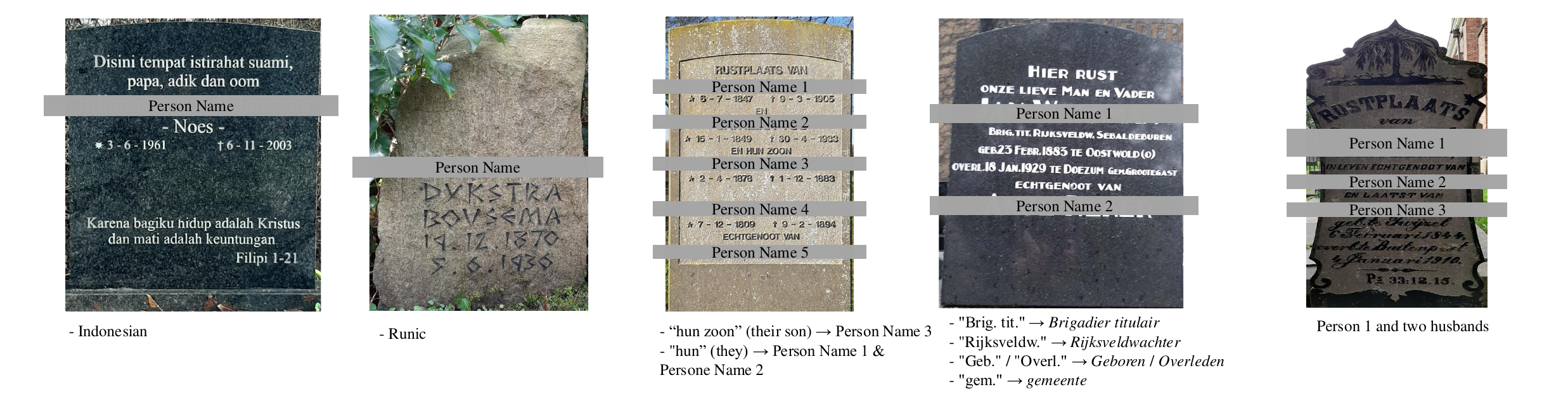}
    \caption{Examples of tombstones with variability and linguistic complexity (left to right): Rare Languages, Rare Font Styles, Coreference, Abbreviations, Multiple Persons.}
    \label{fig:challenge_example}
\end{figure*}

\begin{figure*}[h]
    \centering
    \includegraphics[width=\linewidth]{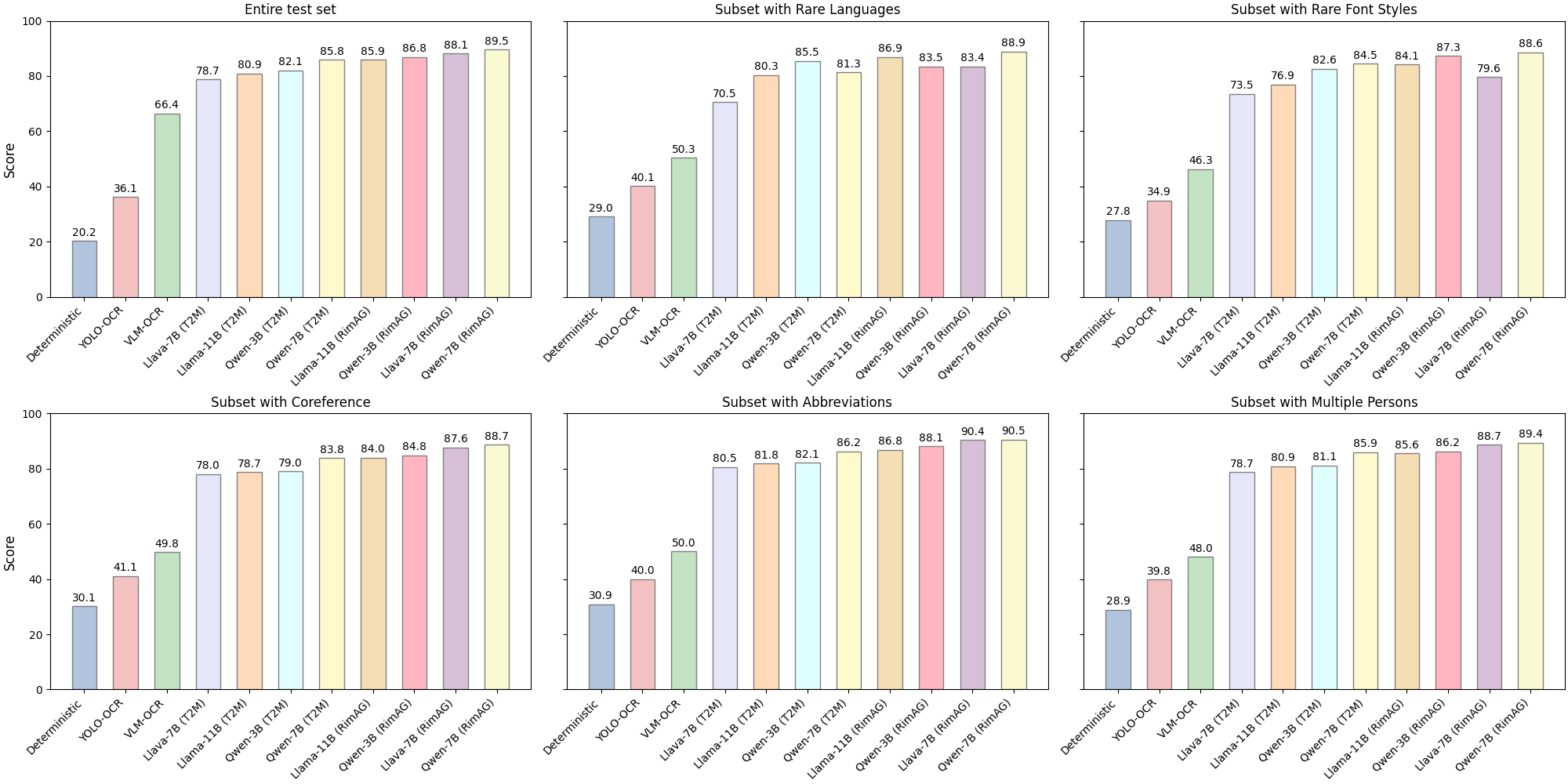}
    \caption{Smatch scores of baselines and T2M on the subsets from five different dimensions.}
    \label{fig:linguistic_performance}
\end{figure*}

\subsection{Challenges in Tombstone Inscriptions}

Tombstone inscriptions are highly diverse, offering cultural insights but posing challenges for automated parsing. Figure~\ref{fig:challenge_example} shows examples varying in style, material, and layout. We identify three key challenges:

\begin{itemize}

    \item Variability: Tombstone inscriptions exhibit significant variation in languages and font styles.

    \item Linguistic Complexity: Inscriptions often include complex linguistic phenomena, such as ambiguous abbreviations, inconsistent date formats and long-distance coreference.

    \item External Knowledge: Semantic interpretation requires linking location names to gazetteers, occupation names to standard database encodings, and concepts to standard ontologies.
    
\end{itemize}

While RAG techniques address the third challenge effectively (as shown in Sections~\ref{sec:methods} and~\ref{sec:exp}), we focus here on evaluating the first two via five curated subsets (Figure~\ref{fig:linguistic_performance}):

\begin{itemize}
    \item \textbf{Rare Languages:} Inscriptions written in languages other than Dutch or in mixed-language formats.
    \item \textbf{Rare Font Styles:} Inscriptions employing non-serif or unconventional fonts.
    \item \textbf{Coreference:} Inscriptions containing long-distance coreferential expressions.
    \item \textbf{Abbreviations:} Inscriptions with shortened or initialed names (e.g., place or occupation names).
    \item \textbf{Multiple Persons:} Inscriptions that reference more than one individual.
\end{itemize}

\begin{figure*}
    \centering
    \includegraphics[width=\linewidth]{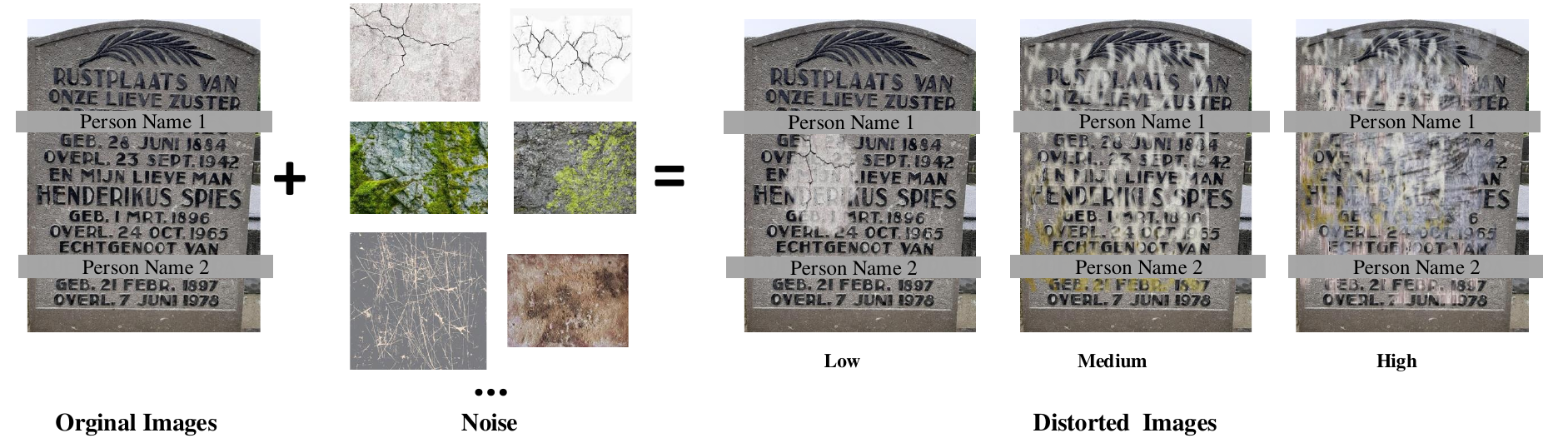}
    \caption{Image fusion for generating noised tombstones. Person names have been blurred to preserve privacy.}
    \label{fig:noise}
\end{figure*}

Figure~\ref{fig:linguistic_performance} shows that our VLM-based framework consistently outperforms all baselines across these five dimensions. In particular, RimAG achieves substantial gains: for example, on the "Abbreviations" subset, Qwen2.5-VL-7B (RimAG) reaches a Smatch score of 90.5, outperforming both standard T2M (82.1) and OCR-based methods (30.9). Similar trends are observed for the "Coreference" subset, where RimAG surpasses 88, while T2M models stay around 83–84, and OCR-based approaches fall below 50. In "Rare Languages" and "Rare Font Styles" subsets, RimAG maintains scores above 88, demonstrating resilience to visual variability. By contrast, deterministic and OCR baselines fall significantly short, with scores as low as 29.0 and 27.8. 

These findings suggest that VLMs possess both expressive and robust capabilities in understanding visual and linguistic inputs, and that RAG techniques remain effective even when models are faced with challenging inscriptions.

\subsection{Challenges of Tombstone Images}

\begin{figure}
    \centering
    \includegraphics[width=\linewidth]{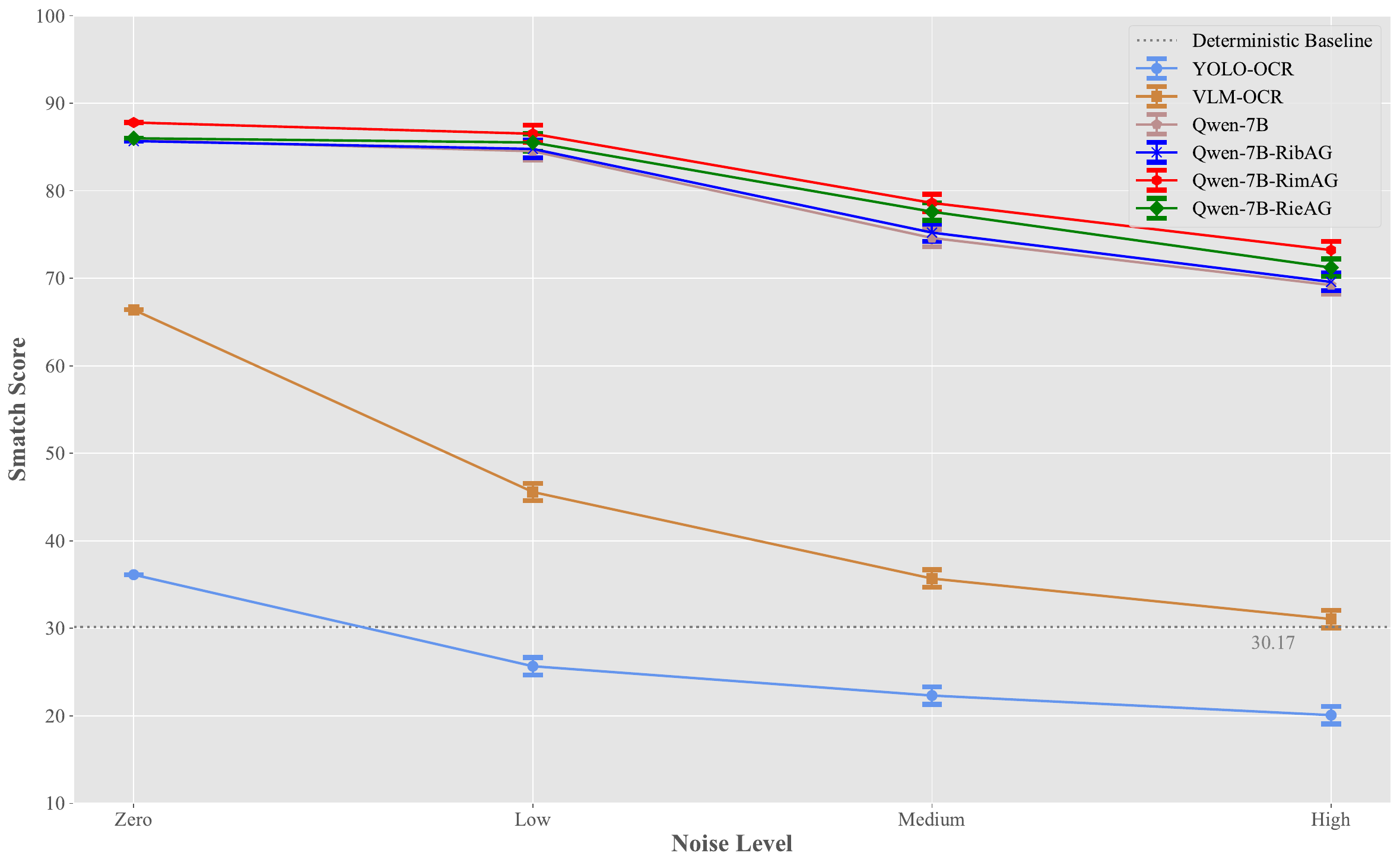}
    \caption{Smatch scores of fine-tuned models on four different noise level (zero, low, medium and high). We only show the best performing augmented framework (Qwen2.5-VL-7B related) to avoid redundancy.}
    \label{fig:noise_result}
\end{figure}

In real-world scenarios, tombstones may exhibit defects such as cracks, erosion, or other damage that affect legibility. While our annotated dataset mainly includes clear inscriptions, we simulate defective tombstones to evaluate robustness. We employ an image fusion technique via alpha blending (see Figure~\ref{fig:noise}), where original images are blended with artificially generated damage. Noise levels are parameterized into low, medium, and high settings (detailed in our supplemental materials). 

Figure~\ref{fig:noise_result} illustrates the performance degradation across four different noise levels. As expected, all models experience a decline in Smatch scores with increasing noise. However, the T2M models exhibit notable robustness; for example, under high noise conditions, Llava maintains a score above 60 while Qwen2.5-VL-7B consistently scores above 70. In contrast, the OCR-based baselines perform at or below the deterministic baseline, highlighting their insufficient robustness in noisy situation. 

Furthermore, retrieval augmentation continues to yield improvements over the base T2M under these challenging conditions, emphasizing the resilience of RAG in tombstone parsing. However, the gains from RibAG are minimal, which may because of the suboptimal performance of OCR when introduce noises. RimAG and RieAG demonstrate substantially greater robustness, achieving improvements comparable to those observed without noise. Overall, our T2M framework exhibits impressive robustness, which is of great significance for its real-world applications.

\section{Conclusion}

In this paper, we introduced the T2M framework combining Vision-Language Models (VLMs) with Retrieval-Augmented Generation (RAG) for the digitization and semantic parsing of tombstone inscriptions. Addressing our first question, we confirmed that VLMs substantially outperform OCR-based methods. Regarding our second question, integrating RAG successfully bridged the critical knowledge gap, enabling integrations of externally dependent entities. Furthermore, the results of the models on tombstone with variability, linguistic complexity and noise confirms that the T2M framework keeps robust performance under challenging conditions, underscoring its practical applicability. 

Despite strong performance, our framework has several limitations. The dataset is mainly Dutch with uniform layouts, limiting generalizability. Simulated degradation may not fully reflect real-world damage such as erosion or occlusion. The RAG component also depends on structured resources, which may be unavailable in certain contexts.

Future work includes expanding language and layout diversity, exploring end-to-end integration of retrieval, and adapting the T2M framework to other heritage artifacts like memorial plaques and archival inscriptions.

\bibliographystyle{ACM-Reference-Format}
\bibliography{sample-base}

\end{document}